\documentclass{llncs}
\usepackage{makeidx}  
\usepackage{graphicx}
\usepackage{amssymb}
\usepackage{hyperref}
\usepackage{xcolor}
\usepackage{wrapfig}
\usepackage{silence}
\usepackage[nice]{nicefrac}
\begin{document}
\frontmatter          
\pagestyle{headings}  
\addtocmark{Hamiltonian Mechanics} 
\title{Are Saddles Good Enough for Deep Learning?}

\titlerunning{saddles in deep learning}  
%
\author{Adepu Ravi Sankar \inst{1} \and Vineeth N Balasubramanian \inst{1}}
\authorrunning{Adepu Ravi Sankar et al} 
%
%
\institute{Indian Institute of Technology Hyderabad, Medak, India\\
\email{cs14resch11001@iith.ac.in, vineethnb@iith.ac.in}}

\maketitle              

\begin{abstract}
Recent years have seen a growing interest in understanding deep neural networks from an optimization perspective. It is understood now that converging to low-cost local minima is sufficient for such models to become effective in practice. However, in this work, we propose a new hypothesis based on recent theoretical findings and empirical studies that deep neural network models actually converge to saddle points with high degeneracy. Our findings from this work are new, and can have a significant impact on the development of gradient descent based methods for training deep networks. We validated our hypotheses using an extensive experimental evaluation on standard datasets such as MNIST and CIFAR-10, and also showed that recent efforts that attempt to escape saddles finally converge to saddles with high degeneracy, which we define as `good saddles'. We also verified the famous Wigner's Semicircle Law in our experimental results.

\end{abstract}

\section{Introduction}
Understanding deep neural networks from an optimization perspective has emerged as an active area of study owing to the great success of deep learning in recent years. Given the non-convex nature of the loss function in deep networks, it is difficult for one to tell where the traditional gradient descent algorithm converged to. It is popularly understood today that deep networks converge to local minima \cite{goodfellow2016deep}. 
Historically, practitioners of neural networks have always considered converging to local minima to be one of the major concerns in neural network training. However, recent work in the last 2-3 years is gradually clarifying this perception, and hypothesizing that converging to global minima is not as essential, but converging to local minima with low cost function values is more important \cite{DBLP:conf/aistats/ChoromanskaHMAL15,Dauphin:2014:IAS:2969033.2969154,goodfellow2014qualitatively,saxe2013exact}. Further, more recent work in \cite{NIPS2016_6112} has shown that all local minima are as good as global minima in very large parameter spaces such as in deep neural networks, and that other critical points are saddle points. At the same time, there have been recent methods such as \cite{Dauphin:2014:IAS:2969033.2969154} and \cite{DBLP:conf/colt/GeHJY15} that attempt to escape saddle points while training. These recent developments in understanding how deep neural networks are trained are also emphasized by the fact that understanding of optimization methods such as gradient descent in low-dimensional spaces does not necessarily translate to very large dimensions. These issues raise very important and pertinent questions, as to where do deep neural network models converge at all. Considering the overwhelming presence of saddle points in very high-dimensional spaces, do neural networks converge to low-cost local minima? Do they instead converge at saddle points? If so, is one saddle point better than the other? Or more importantly, are saddles good enough for deep learning?

The pursuit to answering any of these questions is limited by the fact that characterizing the nature of critical points in high-dimensional spaces is intractable. Also, there has been very limited work in understanding saddle points while training deep networks (Section \ref{sec_saddles_bgd} discusses these efforts). However, this is an important question to answer, and can help provide important insights into the model solutions obtained in deep learning, as well as to find novel methods to converge to `better' solutions more quickly. 

In this work, we attempt the study and characterization of saddle points in deep learning. In particular, we propose the following hypotheses which we study and analyze:
\begin{itemize}
    \item Deep neural network models converge at \textit{degenerate} saddle points (saddles with zero eigenvalues of Hessian).
    \item `Good' saddle points are  often good enough for neural network training.
\end{itemize}
We show empirically that convergence to a `good' saddle (described in Section \ref{sec_escaping_saddles}) suffices to achieve convergence in practically useful deep neural networks (DNNs). To the best of our knowledge, this is the first such effort in characterizing the convergence points of DNN models.

The importance of the contributions in this work lies in clarifying the understanding of how DNNs converge. For a few decades now, it has been understood that DNNs converge to local minima. This work brings a fresh perspective to this understanding by claiming that DNNs actually converge to saddle points. This understanding could fundamentally influence the design of gradient descent algorithms used to train deep networks, and can result in methods that make training more efficient. The understanding derived herein can provide a new dimension to the way we look at Stochastic Gradient Descent (SGD) and its variants, especially with respect to the degeneracy of the parameter space during training. 
Existing theoretical work (such as \cite{DBLP:conf/colt/GeHJY15}) do not consider degeneracy in their analysis, and assume the existence of a strict saddle in their methods. This assumption however does not hold in practice as we show in this work. This work can provide momentum for new theoretical work that takes into account the degeneracy of the saddle points encountered while training DNNs.

The remainder of this paper is organized as follows. Section \ref{sec_saddles_bgd} discusses the background literature and motivation for this work. Section \ref{sec_saddles} proposes the hypothesis that DNN models converge to saddle points and studies this. Section \ref{sec_characterizing_saddles} further characterizes the nature of the saddle that each DNN model converges to. Section \ref{sec_escaping_saddles} then studies what existing methods that attempt to escape saddles actually achieve. We then present analysis of our work in different network settings in Section \ref{sec_discussion}, and conclude with pointers to future work in Section \ref{sec_conclusions}.

\section{Saddles in Deep Networks: Background and Motivation}
\label{sec_saddles_bgd}
It is known that the presence of multiple symmetries in the parameter space is directly correlated to the proliferation of saddle points on the error surface in DNNs. Two common symmetries noted in recent work \cite{badrinarayanan2015understanding}\cite{Dauphin:2014:IAS:2969033.2969154} are \textit{scaling symmetries} and \textit{permutation symmetries}. The presence of scaling symmetries can be explained as follows. Firstly, given $W_1$ and $W_2$ as weight matrices of neighboring layers, scaling $W_1$ by $\alpha$ and $W_2$ by $1/\alpha$ leaves the loss function unchanged. This is particularly relevant when using ReLu activations or linear networks. When batch-norm layers are used in the network, any random scaling of the weight matrices will result in the same symmetry issue. It is known that such symmetries lead to the presence of multiple saddles on the error surface \cite{DBLP:conf/aistats/ChoromanskaHMAL15}. Permutation symmetries occur when the order of hidden units are permuted, and the respective connected weights make the output of a neural network invariant to the input.  These symmetries also cause the Hessian to be degenerate at the critical points \cite{AnandkumarG16}.

 Over the last few years, we have seen a growing attention in the community to the issue of saddle points while training deep networks. Pascanu et al. \cite{pascanu2014saddle} as well as Dauphin et al. \cite{Dauphin:2014:IAS:2969033.2969154} challenged the popular claim that local minima are the main concern while training deep networks. They instead highlighted that while local minima are a concern in low-dimensional parameter spaces, all local minima in high-dimensional spaces have reasonably low costs, and the concern is instead replaced by the presence of saddle points that slow down training. Choromanska et al. \cite{DBLP:conf/aistats/ChoromanskaHMAL15} corroborated this claim by showing that low-index critical points of large models concentrate in a band just above the global minimum (as against critical points of small models that can converge to high-cost local minima). They, in fact, showed that it may not be essential to look for the global minimum, since that may indicate overfitting. These observations have led to a few focused efforts over the last two years on escaping and analyzing saddle points (Ge et al. \cite{DBLP:conf/colt/GeHJY15}, Anandkumar and Ge \cite{AnandkumarG16}, Kawaguchi \cite{NIPS2016_6112}, Lee et al. \cite{DBLP:conf/colt/LeeSJR16}, Hardt et al. \cite{DBLP:conf/icml/HardtRS16}). We introduce some definitions before discussing these methods.
 
\begin{definition}{\textbf{Critical Points.}}
\label{def_critical_points}
Consider a smooth function $f: \mathbb{R}^n \rightarrow \mathbb{R}$. $x$ is a \textit{critical point} iff $\nabla f(x) = 0$. The \textit{critical points} are further classified by considering the \textit{Hessian} $\nabla^2 f(x)$ of $f$ at $x$, :
 \begin{itemize}
     \item If all eigenvalues of $\nabla^2 f(x)$ are positive, critical point $x$ is a \textit{local minimum}.
     \item If all eigenvalues of $\nabla^2 f(x)$ are negative, critical point $x$ is a \textit{local maximum}.
     \item If eigenvalues of $\nabla^2 f(x)$ are both positive and negative, critical point $x$ is a \textit{min-max saddle point}, i.e. if we restrict $f$ to the subspace spanned by the positive (negative) eigenvalues, the saddle point is a local maximum (minimum) of this function.
     \item If there are zero eigenvalues of $\nabla^2 f(x)$ at a saddle point $x$, $x$ is called a \textit{degenerate saddle}. (We define degeneracy of a saddle as the number of zero eigenvalues of the Hessian, $\nabla^2 f(x)$, at $x$.) 
 \end{itemize}
\end{definition}
It is evident from this standard definition that saddle points cover a broad range of scenarios in high-dimensional spaces, depending on the number of positive or negative (or zero) eigenvalues. We note that there is no further standard categorization of saddle points. In order to study saddle points in a constrained setting, Ge et al. \cite{DBLP:conf/colt/GeHJY15} introduced the concept of a ``\textit{strict saddle}", which is defined as below.

\begin{definition}{\textbf{Strict Saddle.}}
\label{strict_saddle_defn}
Given a smooth function $f: \mathbb{R}^n \rightarrow \mathbb{R}$ and a critical point $x$, $x$ is called a $(\alpha,\gamma,\epsilon,\delta)$-\textit{strict saddle}, if one of the following is true:
\begin{itemize}
    \item $\nabla f(x) \geq \epsilon$
    \item $\lambda_{\min} \big(\nabla^2 f(x)\big) \leq - \gamma $
    \item There is a local minimum $w^*$ such that $||w-w^*|| \leq \delta$, and the function $f(w^{*})$ restricted to $2\delta$ neighborhood of $w^*$ (i.e. $||w-w^*|| \leq 2\delta$) is $\alpha$-strongly convex.\hfill 
\end{itemize} 
\end{definition}

There have been disparate efforts to address the issue of saddle points in DNNs, since there is no single solution to this problem. In early work, the existence of saddles in a single hidden layer MLP was shown by Baldi and Hornik \cite{Baldi:1989:NNP:70359.70362}. Pascanu et al. \cite{pascanu2014saddle} and Dauphin et al. \cite{Dauphin:2014:IAS:2969033.2969154} recently identified the problem of saddle points while training deep networks, and proposed a `saddle-free' Newton method to help the optimization method escape saddle points during training. In particular, they proposed a second-order trust region-based method \cite{wright1999numerical} that uses the curvature of the function (as obtained by the Hessian) to define the trust region. Ge et al. \cite{DBLP:conf/colt/GeHJY15} proposed a simple approach called noisy-Stochastic Gradient Descent (\textit{noisy-SGD}), which can escape \textit{strict saddles} in polynomial time. In \textit{noisy-SGD}, a small amount of random noise is added to the calculated gradient at every iteration. The authors claim that this noise helps escape strict saddles. This is easily understood intuitively using the definition of a strict saddle, which requires a strong negative curvature in every direction at the critical point. Hence, adding random noise to the gradient helps SGD escape such saddles. Although this work has theoretical guarantees, there were no empirical studies to validate their claims. Besides, their assumption of having a strong negative curvature in all directions (and thus, no zero eigenvalue in any direction) at critical points (please see Definition \ref{strict_saddle_defn}) is unrealistic for high-dimensional parameter spaces such as in deep neural networks (as we shown in Section \ref{sec_saddles}). 

More recently, Kawaguchi \cite{NIPS2016_6112} showed that in deep neural networks with any depth and width, every local minimum is a global minimum. They also showed that every critical point that is not a global minimum is a saddle point. This is a significant result, and questions the very motive of methods that try to escape saddles, and asks the question: what do DNNs indeed converge to then? In an even more recent work, Jin et al. \cite{jin2017arxiv} showed that a perturbed form of gradient descent always converged to a second-order stationary point in a number of iterations that is `\textit{dimension-free}'. In particular, they state that when all saddle points are non-degenerate, all second-order stationary points are local minima, and the proposed perturbed gradient descent method escapes all saddle points to converge to such local minima.

From these recent efforts, two observations stand out: (i) There continues to be a gap between theory and practice of optimization results in the context of DNNs. Assumptions that are made for proving theoretical results don't hold in practice in such cases. While most of these recent efforts \cite{DBLP:conf/colt/GeHJY15,jin2017arxiv,NIPS2016_6112} show interesting theoretical results, there is very limited validation of these results from empirical studies (optimization results in low-dimensional spaces, unfortunately, don't translate to high-dimensional spaces); and (ii) All the aforementioned recent efforts make the assumption that saddle points in these models are non-degenerate (i.e. there are no eigenvalues of Hessian with value zero at the critical point). This assumption is unrealistic in DNN models, where the parameter space is of the order of millions. Recent related efforts such as Lee et al. \cite{DBLP:conf/colt/LeeSJR16} and Hardt et al. \cite{DBLP:conf/icml/HardtRS16} also provide new insights into convergence properties of gradient descent on non-convex error surfaces, but assume the saddles to be non-degenerate. 

\begin{figure}
\begin{minipage}[b]{.5\linewidth}
\centering
        \includegraphics[scale=0.3]{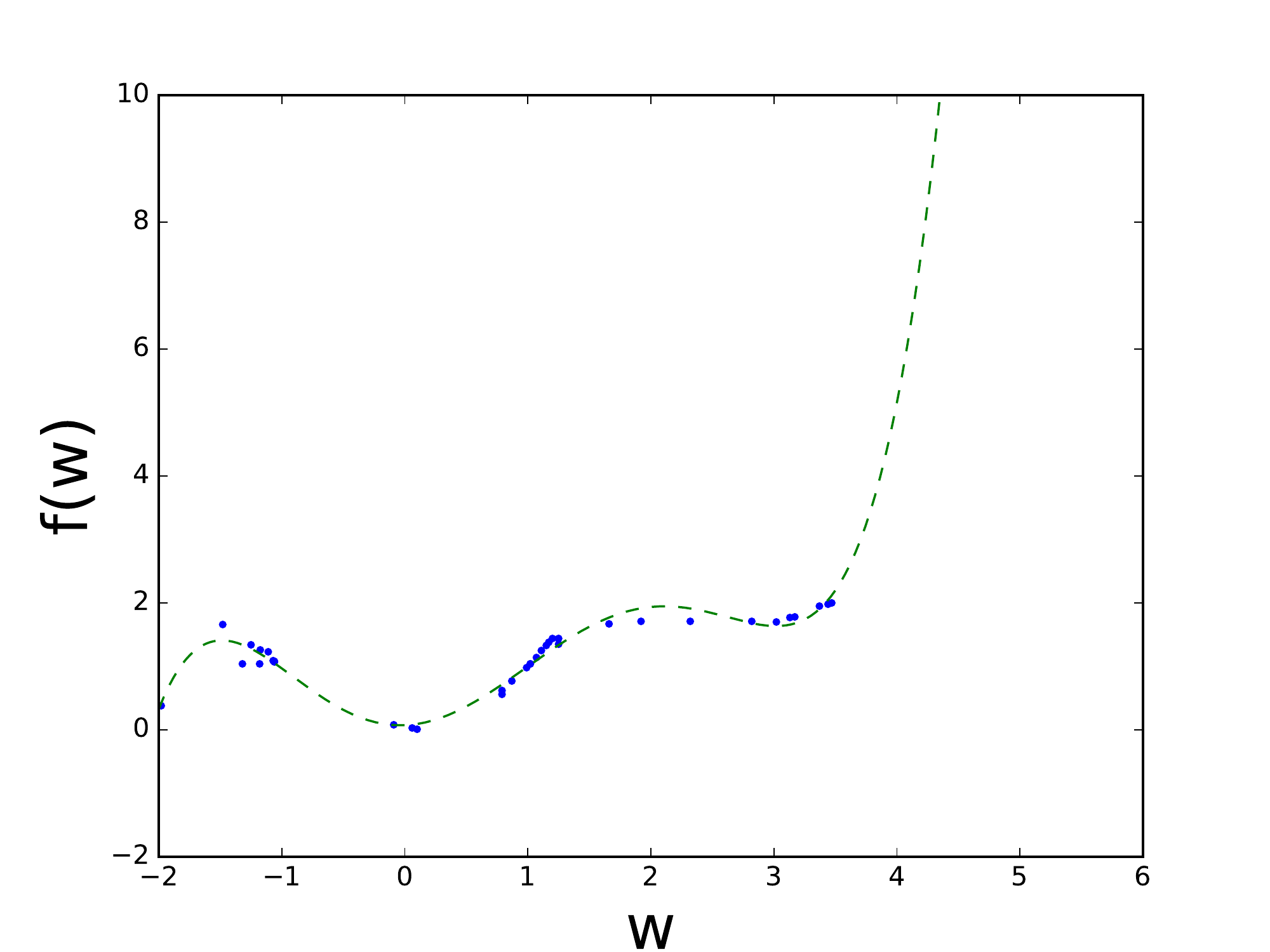}
\end{minipage}%
\begin{minipage}[b]{.5\linewidth}
\centering
        \includegraphics[scale=0.45]{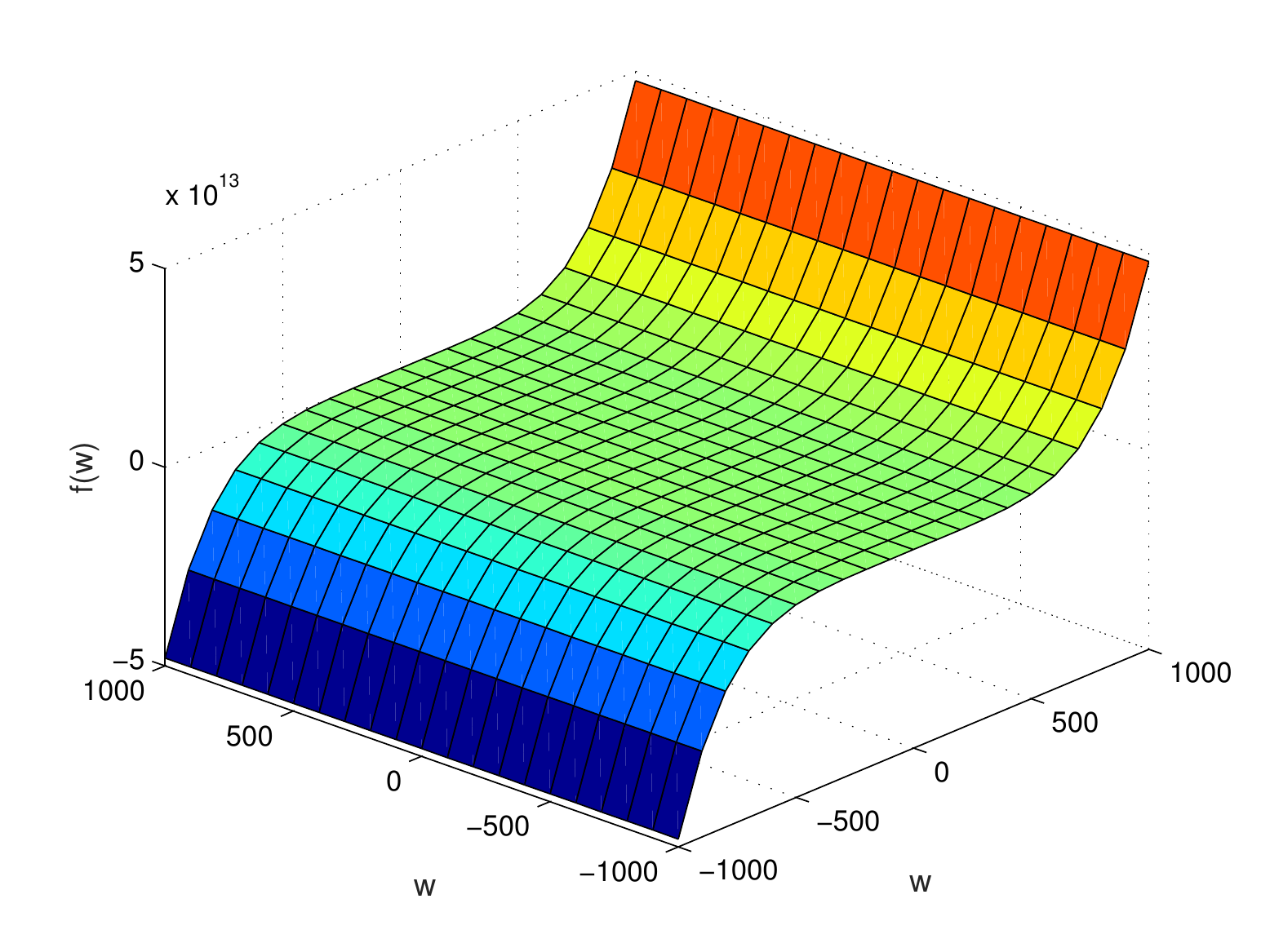}
\end{minipage}
    \caption{A toy example of degenerate saddle}
    \label{example}
\end{figure}

Figure \ref{example} explains the issue of a degenerate saddle using a toy example. Figure \ref{example}a represents the loss function $f(w)$ with a flat plateau in the $w$-interval (2,3). Figure \ref{example}b provides a mesh view of the same function. Although the global minimum of $f$ occurs at (0,0), the traditional gradient descent algorithm converges to the \textit{degenerate saddle} (flat plateau) if the algorithm is initialized at any point above (4,2). Such saddles are a serious concern and it is very difficult for either first-order or second-order methods to escape them. 

A classical work in machine learning by Watanabe in \cite{4233934} states that almost all learning machines are singular where the Fisher information matrix (Generalized Hessian) has zero eigenvalues. These observations highlight the need to study the problem of saddle points in the particular context of high-dimensional DNNs. In addition, Kawaguchi's \cite{NIPS2016_6112} recent claim that any critical point that is not a global minimum is a saddle point seems to contend with the other methods' claim of `escaping saddles' and converging to local minima during DNN training. Hence, in this work, we seek to study these conflicting narratives by characterizing the saddles obtained while training DNNs in practice. We specifically ask the question: ``Do DNNs converge to local minima? Or are saddles good enough?". 


\section{Do Deep Neural Networks Converge to Saddles?}
\label{sec_saddles}

Based on our analysis of recent work and our own studies, in this section, we propose a hypothesis that existing deep network models, including those that have been successful in practice, actually converge at saddle points. 

\newtheorem{hypothesis}{Hypothesis}
\begin{hypothesis}
\label{H1}
\textit{Deep neural network models converge at saddle points.}
\end{hypothesis}

It is theoretically guaranteed that a first-order learning algorithm in a non-convex DNN converges to a critical point. But it was always understood that the converged point is a local minimum. We hypothesize that these DNNs converge to saddles. Answering the above question is not easy and straightforward given the high-dimensional nature of the loss function. While there is preliminary theoretical support for this hypothesis in \cite{NIPS2016_6112}, we designed an extensive experimental evaluation setup to study Hypothesis \ref{H1}, which we describe below. 


Our experimental study to evaluate Hypothesis \ref{H1} was carried out on different datasets which are traditionally used in the deep learning community. We use MNIST \footnote{\url{http://yann.lecun.com/exdb/mnist/}} and CIFAR-10\footnote{\url{https://www.cs.toronto.edu/~kriz/CIFAR.html}} in particular, in this work to evaluate our hypothesis. Both these datasets have been long studied in the community, and the validation of our hypothesis on these datasets has a significant impact. MNIST is a database of handwritten digits from 0 to 9 in grayscale, and each image is of size $28 \times 28$. The training set consists of 60,000 images and test set of 10,000 images. CIFAR-10 is a color image dataset comprising 10 classes with a training set size of 50,000 and test set size of 10,000. The dataset consist of five training batches and one test batch, each with randomly chosen 10,000 images. The source code to reproduce all our results can be found at \url{https://github.com/ravisankaradepu/degenerate_saddle}. 

We train different DNNs on these datasets, and compute the eigenvalues of the Hessian at every iteration. The presence of positive and negative eigenvalues of the Hessian at convergence helps us identify if the corresponding critical point is a saddle. Although the evaluation criteria seems rather trivial, it is difficult in practice (and is also the reason why second-order methods are not used often while training DNNs). The explicit computation of the Hessian matrix is a daunting task and is impossible for larger networks. Hence, the choice of architectures in our experiments was restricted by the capability of the available computational infrastructure in calculating the Hessian\footnote{We used Nvidia's Tesla K20 GPUs for all our experiments}. However, as evident from our experimental results later in this section, the chosen architectures yield results that are comparable to the best results obtained on these datasets. In particular, we trained a Multi Layer Perceptron (MLP) with a single hidden layer to train both MNIST and CIFAR-10. To train MNIST, we used 784 input neurons, 25 hidden neurons with ReLu activations, and 10 output neurons with softmax activations ($784 \times 25 \times 10$) having a cross-entropy loss function. To train CIFAR-10, we used a slightly different procedure. We found that using a similar single-layer MLP for CIFAR-10 does not achieve performance comparable to state-of-the-art on CIFAR-10\footnote{\url{http://rodrigob.github.io/are_we_there_yet/build/classification_datasets_results.html}}. Hence, we instead passed images from CIFAR-10 through a model which has known good performance on the dataset (Wide ResNets \cite{Zagoruyko2016WRN}), and used the features of the penultimate layer (256 dimensions) as the input to a single-layer MLP. Thus, for CIFAR-10, we finally have an MLP with 256 input neurons, 25 hidden neurons with ReLu activations, and 10 output neurons with softmax activations ($256 \times 25 \times 10$) having the cross-entropy loss function. We terminated the learning of all our experiments when the the error difference between two consecutive epochs was less than $10e^{-4}$. The weights of these networks were initialized randomly from a normal distribution. We ran 8 separate trials for all the experiments with this work, and we report the mean and standard deviation of the results across these trials.
The experiments were carried out using different variants of gradient descent methods viz., Momentum \cite{Sutskever:2013:IIM:3042817.3043064}, Adam \cite{DBLP:journals/corr/KingmaB14} and Adagrad \cite{Duchi:2011:ASM:1953048.2021068}. In the momentum method, we used Nesterov momentum with parameter 0.9 (commonly advised by practitioners). In Adam, the parameters $\epsilon$, $\beta_1$, $\beta_2$ are set to $1\mathrm{e}{-8}$, 0.9 ,0.999 respectively (as recommended in \cite{DBLP:journals/corr/KingmaB14}). The results of our experiments on the MNIST dataset is shown in Table \ref{mnist_saddle}, while those of CIFAR-10 are shown in Table \ref{CIFAR_saddle}. Evidently, a significant number of eigenvalues are negative (along with presence of positive eigenvalues) at convergence for both datasets, validating our hypothesis that these deep networks indeed converge to a saddle point.

\begin{table}[h]
\centering
\begin{tabular}{|l|c|c|c|}
\hline
\textbf{METHOD} & \textbf{\shortstack{Positive eigenvalues}} & \textbf{\shortstack{Negative eigenvalues}} & \textbf{Accuracy}\\ \hline \hline
\textbf{Momentum} & 15877 $\pm$ 17  & 2383 $\pm$ 16 &  96.35 $\pm$ 0.02 \% \\ \hline
\textbf{Adam}  & 15958 $\pm$ 80 & 2274 $\pm$ 80 &  96 $\pm$ 04 \% \\ \hline
\textbf{Adagrad} & 15136 $\pm$ 23.76 & 3123 $\pm$ 23.79 & 95 $\pm$ 0.06 \%  \\ \hline \hline
\end{tabular}
\caption{\textbf{MNIST:} Statistics of +ve and -ve eigenvalues of the Hessian at convergence of various gradient descent methods trained on MNIST dataset using a network architecture $784 \times 25 \times 10$. The dimension of Hessian is $19885 \times 19885$, including the biases at each layer.(We explain why the sum of +ve and -ve eigenvalues do not add up to the dimension of the Hessian in the next section.)}
\vspace{-20pt}
\label{mnist_saddle}
\end{table}

\begin{table}[h]
\centering
\vspace*{-\baselineskip}
\begin{tabular}{|l|c|c|c|}
\hline \hline
\textbf{METHOD}         & \textbf{Positive eigenvalues} & \textbf{Negative eigenvalues} & \textbf{Accuracy}\\ \hline \hline
\textbf{Momentum} & 6063 $\pm$ 164 & 410 $\pm$ 86 & 95.02 $\pm$ 0.04\% \\ \hline
\textbf{Adam} & 4701 $\pm$ 314 & 1515 $\pm$ 223 &96.04 $\pm$ 0.02\% \\ \hline 
\textbf{Adagrad}  & 6242 $\pm$ 114 & 272 $\pm$ 44 & 94.94 $\pm$ 0.064\% \\ \hline \hline
\end{tabular}
\caption{\textbf{CIFAR-10:} Statistics of +ve and -ve eigenvalues of the Hessian at convergence of various gradient descent methods trained on CIFAR-10 dataset using a network architecture $256 \times 25 \times 10$. The dimension of Hessian is $6685 \times 6685$, including the biases at each layer. (Note that due to the use of features from Wide Resnets, a smaller network than what we used for MNIST sufficed to achieve state-of-the-art results here. Also, we explain why the sum of +ve and -ve eigenvalues do not add up to the dimension of the Hessian in the next section.)}
\vspace{-10pt}
\label{CIFAR_saddle}
\end{table}

While state-of-the-art models provide an accuracy of $\approx 99\%$ on MNIST, such models use deeper and wider networks, for example, with two hidden layers of 500 neurons in each layer, which totals to around 1.2 million weights. It is intractable to validate this experiment on Hessians of this size. Our network with one hidden layer of 25 neurons has 0.019 million parameters, which is tractable for computing the Hessian, and still provides an accuracy of 96\%, which is comparable to accuracies at convergence of similar networks on MNIST. Table \ref{convergence} (in Section \ref{sec_discussion}) also shows that as the number of hidden neurons in this single-layer MLP is increased, the accuracy increases at convergence along with the increase in model complexity. We chose a network architecture that provides the best performance given the constraint imposed on the model complexity by available computational resources GPU memory of 5 GB allows only a Hessian of size $\approx 20000 \times 20000$ to be computed, which defined the architecture in this work). Our accuracy of 95.02\% on CIFAR-10 is, however, very near to current state-of-the-art results \cite{CIFAR10}, although we use a smaller network. 
We note that the choice of architectures in this work aligns with our focus to understand the behaviour of convergence of gradient descent algorithms in DNNs for highly competitive models.

\section{Characterizing Saddles in DNNs}
\label{sec_characterizing_saddles}

With the validation of our hypothesis that deep networks indeed converge at saddle points, a question that follows is: \textit{are there different categories of saddle points? If so, what kind of saddles do deep networks converge to?} We study this question in this section. While there are various kinds of saddle functions (such as the monkey saddle or the handkerchief saddle), there is no existing literature, to the best of our knowledge, that provides a comprehensive categorization of the kinds of saddle points that occur in practice. Hence, we summarize below the different types of saddles discussed hitherto:
\begin{itemize}
\item \textit{Traditional Saddles:} Corresponding Hessian has positive and negative eigenvalues
\item \textit{Degenerate Saddles:} Corresponding Hessian has zero eigenvalues in addition to positive and negative eigenvalues
\item \textit{Strict saddles:} Corresponding Hessian has only positive and negative eigenvalues, with negative eigenvalues bounded away from zero
\end{itemize}



We hypothesize that in addition to converging to a saddle, deep neural network models that are successful in practice today converge to \textit{degenerate saddles}. While we carry out an empirical validation of this hypothesis in this section, this hypothesis can be understood intuitively too. It is known that first-order learning algorithms stop making further progress when the norm of the gradient becomes zero at a critical point. When the critical point is a saddle, variants of the basic gradient descent algorithms attempt to go to a lower cost critical point by adding momentum or perturbations to the gradient. However, when the the region around the saddle is degenerate or flat (especially when this region of flatness around the critical point is rather large), even such perturbations do not help, thereby letting the network converge.

\begin{hypothesis}
\label{H2}
\textit{Deep neural network models converge at \textit{degenerate saddle points}.}
\end{hypothesis}

To study Hypothesis \ref{H2}, we extended the previous set of experiments from Section \ref{sec_saddles} to capture the \textit{degeneracy} of the saddle that the DNNs converged to. We define degeneracy as the number of zero eigenvalues of the Hessian at the converged point. We now report the degeneracy of the converged points from our experiments in the earlier section ($784 \times 25 \times 10$ for MNIST and $256 \times 25 \times 10$ for CIFAR-10) in Table \ref{tab_degeneracy}. 



\begin{table}[h]
\centering

\begin{tabular}{|l||l|l|l||l|l|l|}
\hline \hline
\multicolumn{1}{|c||}{\textbf{}}  & \multicolumn{3}{c||}{\textbf{MNIST}}       & \multicolumn{3}{c|}{\textbf{CIFAR-10}}    \\ \hline
\multicolumn{1}{|c||}{\textbf{\shortstack{Gradient\\ Descent \\Method}}} & \multicolumn{1}{c|}{\textbf{\shortstack{Deg}}} & \multicolumn{1}{c|}{\textbf{\shortstack{Negative\\e-values}}} & \multicolumn{1}{c||}{\textbf{Accuracy}} & \multicolumn{1}{c|}{\textbf{\shortstack{Deg}}} & \multicolumn{1}{c|}{\textbf{\shortstack{Negative\\e-values}}} & \multicolumn{1}{c|}{\textbf{Accuracy}} \\ \hline \hline
Momentum & 1620 $\pm$ 8  & 2383 $\pm$ 16 & 96.35 $\pm$ 0.02\% & 208 $\pm$ 19 & 410 $\pm$ 86 &  95.00 $\pm$ 0.04\%  \\ \hline
Adam  &  1625 $\pm$ 2 & 2274 $\pm$ 80 & 96.04 $\pm$ 0.02\%        &    376 $\pm$ 4       &  1515 $\pm$ 223 & 95.04 $\pm$ 0.02\%       \\ \hline
Adagrad & 1625 $\pm$ 6 &3123 $\pm$ 23.79     & 95 $\pm$ 0.06\%    &  170 $\pm$ 22 &   272 $\pm$ 44 & 94.94 $\pm$ 0.06\%      \\ \hline \hline
\end{tabular}
\caption{Degeneracy of various gradient descent based methods on network architectures: Input(784) $\times$ Relu(25) $\times$ Output(10) for MNIST; and Input(256) $\times$ Relu(25) $\times$ Output(10) for CIFAR-10. Total eigenvalues for MNIST = 19885; for CIFAR-10 = 6695, as before. (Deg = degeneracy; e-values = eigenvalues)}

\label{tab_degeneracy}
\end{table}

The results show that a significant number of eigenvalues are zero at convergence for both datasets, while providing a test accuracy on par with state-of-the-art methods. This is observed for all the considered gradient descent methods. Table \ref{tab_degeneracy} indicates that the saddle is flat along upto $\approx 1600$ dimensions in the case of momentum-based gradient descent in MNIST, which highlights the extent of degeneracy in the saddle. The negative eigenvalues are included to show that the critical points are indeed saddles. The flatness of the converged point makes it difficult for the gradient-descent based algorithms to escape the saddles. Table \ref{tab_degeneracy} thus validates Hypothesis \ref{H2} that the converged point is indeed a \textit{degenerate saddle}.



\begin{figure}
    \centering
    \includegraphics[height=5cm, width=10cm]{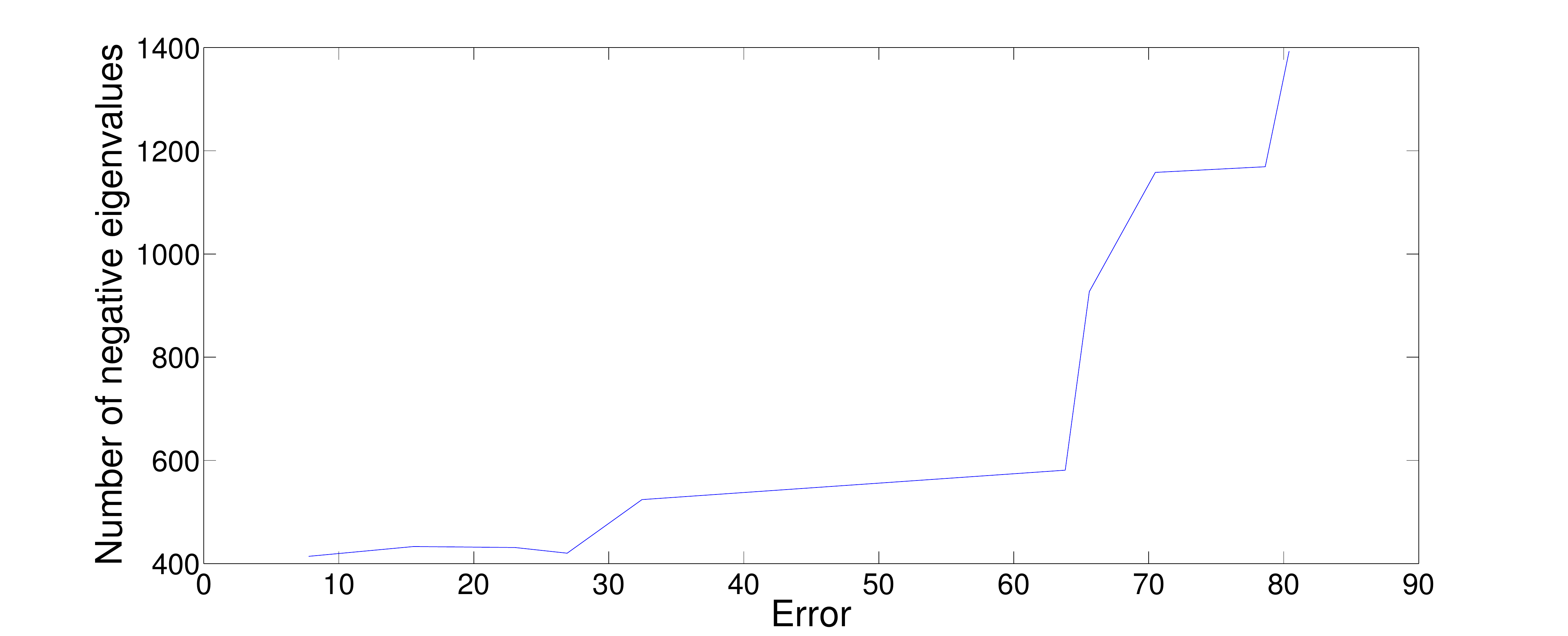}
    \caption{$\epsilon-\alpha$ graph while training a network with $256$(Input) $\times 25$ (ReLu) $\times 10$ (Softmax) on CIFAR-10}
    \label{epsilon-alpha}
\end{figure}

One other way of analyzing the Hessian at critical points is by using $\epsilon$-$\alpha$ plots proposed by Bray et al. \cite{bray:hal-00124320}, where $\alpha$ is the fraction of negative eigenvalues of Hessian at the critical point and $\epsilon$ is the error obtained at the critical point. Bray et al. \cite{bray:hal-00124320} noticed that on the $\epsilon$ vs $\alpha$ plane, the critical points concentrate on a monotonically increasing curve as $\alpha$ ranges from 0 to 1, which implies a very strong correlation between the fraction of negative eigenvalues of the Hessian and the accuracy. Figure \ref{epsilon-alpha} shows the $\epsilon$-$\alpha$ plot for CIFAR-10 trained on the network $256 \times 25 \times 10$. We can observe in the plot that as
the negative eigenvalues on $y$-axis increases, the $\epsilon$ also increases almost monotonically. This observation is understood better along with the famous Wigner's semicircle law ~\cite{wigner_semicircle}, which is stated below.

\begin{theorem}
{\textbf{Wigner's Semicircle Law.}}
Let $V$ be a real symmetric matrix of large order $N$ having random elements $v_{ij}$ that for $i \leq j$ are independently distributed with equal densities, equal second moments $m^2$, and $n^{th}$ moments bounded by constants $B_n$ independent of $i$,$j$ and $N$. 
Further, let $S=S_{\alpha,\beta}(v,N)$ be the number of eigenvalues of $V$ that lie in the interval $(\alpha N^{1/2}, \beta N^{1/2})$ for real $\alpha < \beta$. Then:
\[ \lim_{N \rightarrow \infty} \frac{E(S)}{N} = \frac{1}{2 \pi m^2} \int_{\alpha}^{\beta} \sqrt{4m^2 -x^2} dx\]
\end{theorem}

In simple words, Wigner's law states that for large random symmetric matrices, the distribution of eigenvalues appears like a semicircle with mean at zero (we note that the Hessian is symmetric in our case). Also, for random matrices, as the dimension of the matrix increases, the probability of zero eigenvalues in a random matrix is nearly \nicefrac{1}{2} \cite{wigner_semicircle}. Dauphin et al. \cite{Dauphin:2014:IAS:2969033.2969154} also noted that the semicircle law holds for Hessians obtained in deep learning settings. Now, from the $\epsilon$-$\alpha$ plots, the number of negative eigenvalues decreases with lowering error (i.e. as training progresses). Hence, at convergence, while the number of negative eigenvalues may have reduced over the course of the training, due to Wigner's law, it is only expected that either the number of zero eigenvalues increases or many positive eigenvalues come closer to zero, which directly influences the degeneracy of the saddle. This again supports our Hypothesis \ref{H2} that deep neural network models converge at degenerate saddle points.

\section{Escaping Saddles: What does this mean?}
\label{sec_escaping_saddles}

As discussed in Section \ref{sec_saddles_bgd}, there have been a few methods proposed over the last 2-3 years on escaping saddle points while training DNNs \cite{Dauphin:2014:IAS:2969033.2969154,DBLP:conf/colt/GeHJY15}. The validation of our Hypothesis \ref{H2} in Section \ref{sec_characterizing_saddles} that deep neural networks actually converge to a kind of saddle (\textit{degenerate saddle}) in practice raises two further important questions: (i) \textit{Is there a `good' saddle that one would want a deep neural network to converge to?}; and (ii) \textit{What do methods that propose to escape saddle points achieve?} We answer these questions in this section, starting with the second one.

Dauphin et al. \cite{Dauphin:2014:IAS:2969033.2969154} proposed a saddle-free Newton method to escape saddle points while training DNNs. This method uses Krylov subspace descent, where the loss function is optimized in a lower-dimensional Krylov subspace calculated through Lanczos iteration of the Hessian. More recently, Ge et al. \cite{DBLP:conf/colt/GeHJY15} proposed the noisy-SGD to escape strict saddles (Definition \ref{strict_saddle_defn}), which was followed by Jin et al. \cite{jin2017arxiv}'s perturbed gradient descent to also escape strict saddles. These recent efforts are summarized in Table \ref{tab_saddle_escape_methods}.

\begin{table}[h]
\centering        
\vspace*{-\baselineskip}
\begin{tabular}{|l|l|p{6cm}|}
\hline \hline
\textbf{\shortstack{Type of \\ saddle addressed}} & \textbf{Method} & \textbf{Description}  \\ \hline \hline
Traditional   & Saddle-Free Newton \cite{Dauphin:2014:IAS:2969033.2969154}       &   Uses a Krylov subspace-based approach and a heuristic to rescale the gradient by $\nicefrac{1}{\left| \lambda_i\right|}$  \\ \hline
Strict saddle & Noisy SGD \cite{DBLP:conf/colt/GeHJY15} & Adds a small amount of random uniform noise to the gradient to help escape saddle\\ \hline
Strict saddle & Perturbed GD \cite{jin2017arxiv} & Adds noise sampled from a unit ball to the parameters before performing gradient descent in certain iterations\\ \hline \hline
\end{tabular}
\caption{Summary of methods proposed to escape saddle points in DNNs in recent years}
\vspace{-10pt}
\label{tab_saddle_escape_methods}
\end{table}

\begin{table}[h]
\centering
\vspace*{-\baselineskip}
\begin{tabular}{|l|l|l|l|l|}
\hline \hline
\multicolumn{1}{|c|}{\textbf{}} & \textbf{+ve eigenvalues} & \textbf{Zero eigenvalues} & \multicolumn{1}{c|}{\textbf{-ve eigenvalues}} & \textbf{Accuracy} \\ \hline \hline
\textbf{MNIST}  & 15546 $\pm$ 77  &1625 $\pm$ 3 & 2713 $\pm$ 77  &96.47 $\pm$ 0.14\% \\ \hline
\textbf{CIFAR-10} &  6200 $\pm$ 103  & 51 $\pm$ 6  & 357 $\pm$ 29 & 94.99 $\pm$ 0.03\%   \\ \hline \hline
\end{tabular}
\caption{Study of nature of convergence point on saddle escaping method, \textit{noisy-SGD}, on MNIST and CIFAR-10}
\vspace{-10pt}
\label{tab_noisy_sgd}
\end{table}
In order to understand what these methods achieve in the light of the validation of our Hypothesis \ref{H2} in Section \ref{sec_characterizing_saddles}, we carried out an empirical evaluation by performing a similar set of experiments as we did to prove Hypotheses \ref{H1} and \ref{H2}. We analyzed the nature of the converging point of two of the aforementioned methods in Table \ref{tab_saddle_escape_methods}: Noisy-SGD \cite{DBLP:conf/colt/GeHJY15} and Saddle-free Newton \cite{Dauphin:2014:IAS:2969033.2969154}. As already noted before, the assumption of \textit{strict saddles} made by Ge et al. in \cite{DBLP:conf/colt/GeHJY15} does not hold in practice, but we nevertheless studied their method. We trained the same architectures described in earlier sections: $784 \times 25$(ReLu) $\times 10$ (Softmax) neural network on MNIST, and a $256 \times 25$(ReLu)$\times 10$ (Softmax) neural network on CIFAR-10. Table \ref{tab_noisy_sgd} presents the results of Noisy-SGD at convergence. It is evident from the results that even this method, that attempts to escape saddles, converges to a \textit{degenerate saddle}, while providing performance comparable to state-of-the-art. In case of Saddle-free Newton, the authors' implemented this method on an autoencoder for MNIST as in \cite{Dauphin:2014:IAS:2969033.2969154}. We used an architecture of $784 \times 10 \times 784$ with the dimension of the Hessian being $16474 \times 16474$, and were again constrained by limitations of our computational resources in expanding to a larger architecture. At convergence of the autoencoder (which has the same criteria of the error differing by less than $10e^{-4}$ between consecutive iterations), the degeneracy (number of zero eigenvalues) was found to be $\approx 1625$.

It is evident from the above results that even methods that propose to escape saddle points converge at degenerate saddles, thus validating both our hypotheses \ref{H1} and \ref{H2} in this work. We surmise that these methods only seem to take the gradient descent method from a `poor' saddle to a `better' saddle over the process of training a DNN. This brings us to the first question we raised in the beginning of this section: \textit{what indeed is a good saddle?} While there is no definite answer to define a `good' saddle, we coin a definition of the kind of saddles that may be good enough for DNNs to converge to.

\begin{definition}
{\textbf{$(\kappa,\epsilon,\rho)$-Stable Saddle.}}
\label{def_stable_saddle}
An $\epsilon$-second-order saddle point of $f$ with a $\rho$-Hessian Lipschitz function is a \textit{$(\kappa,\epsilon,\rho)$-stable saddle} of $f$ if  $||\nabla f(x)|| < \epsilon$ and 
 $||$diag$(\Lambda (\nabla^2 f))||_0 \leq \kappa$, where $\kappa < N$ is a positive number with $N$ being the dimension of the Hessian $\nabla^2 f$, and $\Lambda$ is the diagonal matrix with eigenvalues of $\nabla^2 f$.
\end{definition}

In other words, Definition \ref{def_stable_saddle} states that the number of non-zero eigenvalues of the Hessian is bounded by $\kappa < N$, thus making this a degenerate saddle. The condition on the Hessian to be $\rho$-Lipschitz states that: $\forall x_1,x_2$, $ || \nabla^2 f(x_1) - \nabla^2 f(x_2) || \leq \rho || x_1 -x_2|| $. This ensures that the function is well-behaved near the saddle point. While the above definition only specifies the stability of the saddle, it is also important for the saddle to be associated with a low error on the associated learning problem, which we state below.


\begin{definition}{\textbf{$(\kappa,\epsilon,\rho,\delta)$-Good Saddle.}}
\label{def_good_saddle}
A \textit{$(\kappa,\epsilon,\rho)$-stable saddle} of $f$ is also a $(\kappa,\epsilon,\rho,\delta)$-good saddle for $f$ if it provides an upper bound, $\delta$, on the error $f(x) - f(x^*)$, where $x^*$ is the optimal solution.  
\end{definition}

\begin{figure}
\centering
\includegraphics[width=0.7\textwidth]{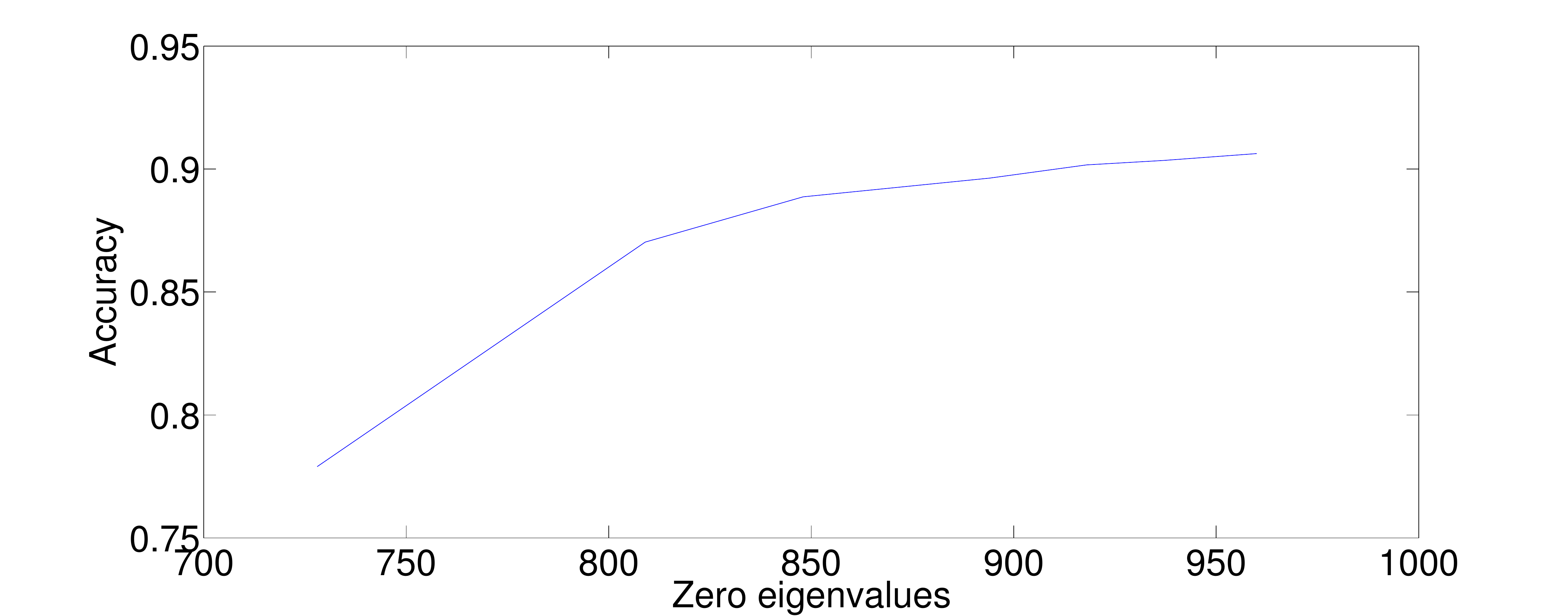}
\caption{Steady increase of zero eigenvalues while training}
\label{fig_deg}
\end{figure}

Evidently, our critical points at convergence in all our results so far satisfy the above definition of a good saddle.
In order to validate our claim of deep networks converging to the above definition of a `good saddle', we also studied how the degeneracy changes at every epoch. We note that this is not an easy experiment, since computing the Hessian at every epoch drastically slows down the training, and is compute-intensive. We hence trained a much smaller network to see this pattern of degeneracy. This result is shown in Figure \ref{fig_deg}, which indicates that the degeneracy (number of zero eigenvalues) of the parameter space increases with every epoch during training. This can also be explained by the fact that the learning algorithm is able to escape saddles with lesser degeneracy initially, but as the degeneracy of the saddle increases, the learning algorithm finds it difficult to escape due to the wide plateau around the converged point.

\section{Discussion}
\label{sec_discussion}

The results discussed so far clearly validate that deep models tend to converge at degenerate saddles. This inference may provide a new dimension to the way researchers approach SGD and similar gradient descent algorithms for training DNNs. Theoretical efforts over the last few years have proposed methods to escape saddles, but with the assumption that the saddles have no zero eigenvalues. As evident through this work, this is not a practical assumption, and this may provide motivation for newer theoretical work that considers the degeneracy of saddles.

In addition to the experiments conducted so far, we also studied the impact of weight initialization and network depth on the hypotheses proposed in this work. Many recent efforts such as \cite{Sutskever:2013:IIM:3042817.3043064} have claimed that good weight initialization helps converge to a better critical point while training DNNs. Table \ref{tab_weight_init} summarizes few weight initialization methods proposed in the recent past.

\begin{table}[h]
\centering
\vspace*{-\baselineskip}
\begin{tabular}{|l|l|}
\hline \hline
\multicolumn{1}{|c|}{\shortstack{Initialization \\ method}} & \multicolumn{1}{c|}{\textbf{Description}} \\ \hline \hline
Xavier's \cite{Glorot10understandingthe}     & $ W \sim U\big{[} - \frac{\sqrt{6}}{\sqrt{n_j+n_{j+1}},},  \frac{\sqrt{6}}{\sqrt{n_j+n_{j+1}},}\big{]}$,  $n_j$ = number of neurons at layer $j$         \\ \hline
He's \cite{He15delvingdeep} &  $ W \sim U\big{[} - \frac{4}{n_j+n_{j+1},},  \frac{4}{n_j+n_{j+1},}\big{]}$                 \\ \hline
Orthogonal                          &        Orthogonal basis used to initialize weight matrix         \\ \hline
Normal (uniform)                            &   Sample weights from normal (uniform) distribution                   \\ \hline  \hline
\end{tabular}
\caption{Weight initialization methods, including recently proposed methods}
\vspace{-30pt}
\label{tab_weight_init}
\end{table}

\subsubsection{Impact of weight initialization on converged point:}
We experimented with various types of initializations to see the degeneracy at the converged point. Table \ref{weight} studies the effect of various successful weight initialization methods on standard SGD with Nesterov momentum. We use the same network architectures of $784 \times 25 \times 10$ for MNIST and $286 \times 25 \times 10$ for CIFAR-10 training, as described in earlier sections. (We only present the degeneracy here for sake of convenience of presentation). Table \ref{weight} convinces us that irrespective of the weight initialization method, the network always converged to a degenerate saddle.

\begin{table}[h]
\centering
\vspace*{-\baselineskip}
\begin{tabular}{|l|l|l|l|l|}
\hline \hline
\multicolumn{1}{|c|}{\textbf{}}       & \multicolumn{2}{c|}{\textbf{MNIST}}       & \multicolumn{2}{c|}{\textbf{CIFAR 10}}    \\ \hline
\multicolumn{1}{|c|}{\textbf{Initialization method}} & \multicolumn{1}{c|}{\textbf{Degeneracy}} & \multicolumn{1}{c|}{\textbf{Accuracy}} & \multicolumn{1}{c|}{\textbf{Degeneracy}} & \multicolumn{1}{c|}{\textbf{Accuracy}} \\ \hline \hline
He's & 1625 & 96.02 $\pm$ 0.1 & 576 $\pm$ 210 & 94.92 $\pm$ 0.09 \\ \hline
Normal & 1625  & 96.15 $\pm$ 0.07 & 423 $\pm$ 111 & 95.02 $\pm$ 0.04 \\ \hline
Xavier's & 1529 $\pm$ 9 & 96.21 $\pm$ 0.2 & 403.5 $\pm$ 113 & 95.01 $\pm$ 0.10\\ \hline
Orthogonal & 1571 $\pm$ 11  & 96.07 $\pm$ 0.03& 542 $\pm$ 218 & 94.99 $\pm$ 0.07 \\ \hline \hline
\end{tabular}
\caption{Study of effect of various initialization methods on SGD convergence. The Hessian in MNIST is $19885 \times 19885$ and in CIFAR-10 is $6695 \times 6695$, as before.}

\label{weight}
\end{table}

\subsubsection{Impact of network depth on converged point:}
Since depth plays a prominent role in the success of DNNs, we also investigated the relationship of network depth on the nature of the converged critical point. The motivation for this study also comes from Wigner's Semicircle law ~\cite{wigner_semicircle}. Wigner proved that for random matrices, as the dimension of the matrix increases, the probability of zero eigenvalues in a random matrix is nearly \nicefrac{1}{2}. Network depth leads to increase in the size of parameters, and hence, the Hessian. 
Table \ref{depth} studies the effect of depth on the degeneracy. We performed experiments till a depth of 4, beyond which it becomes intractable to explicitly compute the Hessian on our current hardware setup. Table \ref{depth} firstly verifies Wigner's Semicircle law in our DNN setup (as discussed in Section \ref{sec_characterizing_saddles}). While we have not explicitly shown the count of positive and negative eigenvalues, these experiments converged to saddles too. The degeneracy of the saddle gradually increases with increasing depth. The number of parameters in a depth-1 MLP for MNIST is 19885 (the architecture described in earlier sections), and the corresponding number of zero eigenvalues was about 1625, with the fraction of zero eigenvalues being 0.08. The number of parameters in depth-2 and depth-3 networks, are 20525 and 21150 respectively, and the corresponding fraction of zero eigenvalues are 0.094 and 0.10 showing the pattern of increasing degeneracy. A similar trend is also observed during the training of CIFAR-10. To the best of our knowledge, this is the first effort to empirically validate Wigner's Semicircle law in deep neural networks.

\begin{table}[h]
\centering
\vspace*{-\baselineskip}
\begin{tabular}{|l|l|l|l|l|l|l|}
\hline
\multicolumn{1}{|c|}{}     & \multicolumn{3}{c|}{\textbf{MNIST}}       & \multicolumn{3}{c|}{\textbf{CIFAR 10}}    \\ \hline
\multicolumn{1}{|c|}{\textbf{Depth}} & \multicolumn{1}{c|}{\textbf{Architecture}} & \multicolumn{1}{c|}{\textbf{Deg}} & \multicolumn{1}{c|}{\textbf{Acc}} & \multicolumn{1}{c|}{\textbf{Architecture}} & \multicolumn{1}{c|}{\textbf{Deg}} & \multicolumn{1}{c|}{\textbf{Acc}} \\ \hline \hline
3      & $784 \times 25 \times 25 \times 10$      & 1847 $\pm$ 63       & 95.97 $\pm$ 0.02    & $256 \times 25 \times 25 \times 10$      & 96.5 $\pm$ 25        & 95.03 $\pm$ 0.07\\ \hline
4      &  $784 \times 25 \times 25 \times 25 \times 10$   & 2119 $\pm$ 17       & 96.06 $\pm$ 0.04    &  $256 \times 25 \times 25 \times 25 \times 10$  &  216 $\pm$ 44&     95.85 $\pm$ 0.08     \\ \hline \hline
\end{tabular}
\caption{Effect of network depth on converged saddles. Deg = Degeneracy; Acc = Accuracy}
\vspace{-10pt}
\label{depth}
\end{table}

As mentioned in Section \ref{sec_saddles}, while the architecture of our network on CIFAR-10 is close to par with state-of-the-art performance, our network architecture for MNIST is relatively lower, and this is because a higher accuracy requires deeper networks, for which it is not tractable to compute the Hessian. 
\begin{table}[h]
\centering
\vspace{-20pt}
\begin{tabular}{|l|l||l|l|}
\hline
\multicolumn{2}{|c||}{\textbf{MNIST}}         & \multicolumn{2}{c|}{\textbf{CIFAR 10}}      \\ \hline
\multicolumn{1}{|c|}{\textbf{Architecture}} & \multicolumn{1}{c||}{\textbf{Accuracy}} & \multicolumn{1}{c|}{\textbf{Architecture}} & \multicolumn{1}{c|}{\textbf{Accuracy}} \\ \hline \hline
$784 \times 500 \times 10$ & 98 $\pm$ 0.09 & $256 \times 500 \times 10$ & 94.98 $\pm$ 0.02   \\ \hline
$784 \times 250 \times 10$ & 97.78 $\pm$ 0.13 & $256 \times 250 \times 10$ & 95.07 $\pm$ 0.08\\ \hline
$784 \times 100 \times 10$ & 97.62 $\pm$ 0.16 & $256 \times 100 \times 10$ & 95.03 $\pm$ 0.02 \\ \hline
$784 \times 50 \times 10$  & 97.03 $\pm$ 0.16 & $256 \times 50 \times 10$ & 95.01 $\pm$ 0.02\\ \hline \hline
\end{tabular}
\caption{Performance of network architectures with increased number of hidden neurons on MNIST and CIFAR-10}
\label{convergence}
\end{table}
We conducted an experiment to show that increasing the number of hidden neurons in the MNIST architecture chosen in this work actually takes the performance of our network close to state-of-the-art. Table \ref{convergence} presents these results. In other words, these results show that MNIST architecture chosen in this work provides the best-in-class performance for a network of this size. We also conducted this experiment on our CIFAR-10 architecture, and observed that the architecture we chose in our work was quite good itself.

\section{Conclusions and Future Work}
\label{sec_conclusions}
In this work, we proposed a new hypothesis that deep neural network models that use gradient descent methods for training  often converge to \textit{degenerate saddle} points. We validated this hypothesis using an experimental evaluation on standard datasets such as MNIST and CIFAR-10. We further studied the nature of convergence points of methods that have been recently proposed for escaping saddles, and found that in these cases too, the models converged to degenerate saddles. Our extensive experiments in this work have provided a fresh perspective to the understanding of the training of deep networks, and can have a direct impact on newer optimization methods proposed to train deep networks.
In future work, we will explore the theoretical implications of our observations, especially as it pertains to results from random matrix theory. We also plan to investigate methods that attempt to escape higher-order saddles (as in \cite{AnandkumarG16}) in the context of the findings of this work.

\section{Acknowledgments}
This work was supported by MHRD-Govt of India and Intel India. We thank them for their generous support. We also thank Yurii Shevchuk (author of neupy library) for his support.

\bibliography{ecml}

\bibliographystyle{splncs03} 

\end{document}